\documentclass[runningheads]{llncs}

\usepackage{eccv}

\usepackage{eccvabbrv}

\usepackage{graphicx}
\usepackage{booktabs}
\usepackage[table]{xcolor}
\usepackage[accsupp]{axessibility}  
\usepackage{multirow}
\usepackage{algorithm}
\usepackage{algpseudocode}
\usepackage{enumitem}
\usepackage{caption}

\usepackage{hyperref}

\usepackage{orcidlink}

\begin{document}

\title{Memory-Supported Synergistic Adaptation \\ for Training-Free Test-Time \\ Medical Image Segmentation} 

\titlerunning{MSSA for Training-Free Test-Time Medical Image Segmentation}

\author{Lingrui Li\inst{1} \and
Nan Pu \inst{2*}  \and
Dong Zhao \inst{3*}\and
Wenjing Li \inst{2} \and \\
Andrew P French \inst{1} \and
Zhun Zhong\inst{2} \and
Xin Chen \inst{1} }

\authorrunning{L.Li et al.}

\institute{School of Computer Science, University of Nottingham, UK \and
School of Computer Science and Information Engineering, Hefei University of Technology, China
\and
Information Systems Technology and Design Pillar, Singapore University of Technology and Design, Singapore
\\
}


\maketitle

\renewcommand{\thefootnote}{}
\footnotetext{* Corresponding author.}
\footnotetext{Author’s Accepted Manuscript. Released under the Creative Commons license: Attribution 4.0 International (CC BY 4.0) https://creativecommons.org/licenses/by/4.0/}

\begin{abstract}
Test-time adaptation (TTA) aims to mitigate distribution shifts by adapting models with unlabeled target data at inference time. While TTA with vision-language models (VLMs) has shown promising results in classification, extending it to medical image segmentation remains challenging. In this setting, the adaptation gains from optimizing on VLM-generated predictions are often outweighed by the degradation to the VLM’s strong pretrained features caused by noisy, update-driven learning, resulting in limited and unstable improvements.
We therefore propose \textbf{Memory-Supported Synergistic Adaptation (MSSA)}, a novel \textbf{training-free} TTA framework for medical image segmentation. Without updating model parameters, MSSA dynamically selects reliable image--text predictions to construct an online memory, uses them as text-guided semantic priors, and couples them with cross-image structural alignment for robust adaptation. Specifically, MSSA consists of (i) a noise-aware memory construction module that filters and stabilizes cross-modal predictions, and (ii) a relevance-driven prototype alignment module that aligns the target sample with structurally consistent memory samples and their reliable predictions to improve adaptation. 
Extensive experiments on multiple medical segmentation benchmarks demonstrate that  MSSA consistently improves VLM-based segmentation models and outperforms existing fine-tuning-based TTA methods by a clear margin, with gains of up to 12.2\% DSC and 11.7\% mIoU. Project Page: https://lingrayy.github.io/MSSA/.

  \keywords{Training-Free Test-Time Adaptation \and Medical Image Segmentation \and Prototype Matching}
\end{abstract}

\section{Introduction}
\begin{figure*}[t]
    \centering
\includegraphics[width=\linewidth]{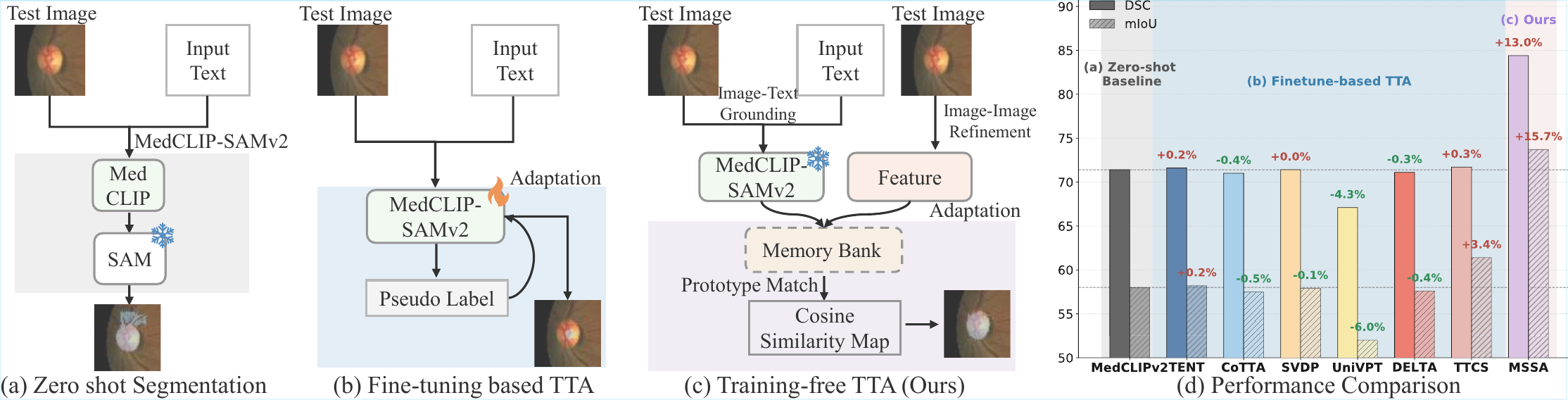}
\captionof{figure}{Comparison of Test-Time Adaptation (TTA) paradigms.
(a) Zero-shot VLM segmentation~\cite{medclipv2} relies on image--text alignment and often suffers from unstable localization under domain shifts.
(b) Fine-tuning-based TTA~\cite{TTCS} updates model parameters using noisy image--text pseudo-labels, leading to performance drift.
(c) Our MSSA achieves training-free adaptation through synergistic image--text grounding and image--image prototype refinement.
(d) Quantitative comparison showing that MSSA achieves consistent improvements while fine-tuning-based TTA methods~\cite{cotta,tent,SVDP,univpt,delta,TTCS} are prone to noise amplification. Improvements are measured relative to Zero Shot in (a).}
\label{fig:motivation}
\end{figure*}

Recent advances in large-scale Vision--Language Models (VLMs) have significantly reshaped the landscape of medical image analysis~\cite{clip,sam,biomedclip}. These VLMs demonstrate strong zero-shot segmentation capabilities by aligning visual and textual embeddings through attention mechanisms, enabling flexible and prompt-driven segmentation~\cite{medclipv1,medclipv2} (see Fig.~\ref{fig:motivation}(a)).
Despite their impressive generalization ability, we observe that foundation models remain vulnerable to domain shifts in real-world medical scenarios. 
When the downstream target data exhibit clear discrepancies from the pre-training distribution, performance degradation becomes inevitable. 
This domain gap limits the practical deployment of VLM-based segmentation systems.

Test-Time Adaptation (TTA)~\cite{cotta,tent,SVDP,MGIPT} aims to mitigate distribution shifts by adapting models using unlabeled target data on-the-fly. 
In segmentation tasks, most TTA methods~\cite{tent} primarily focus on Vision Foundation Models (VFMs), while effective adaptation of Vision–Language Models (VLMs) remains largely underexplored.
TTCS~\cite{TTCS} represents an early attempt to extend TTA to medical VLM-based segmentation (see Fig.~\ref{fig:motivation}(b)); however, the performance gains are marginal.
Moreover, our empirical study shows that when representative training-based TTA methods designed for VFMs~\cite{delta,SVDP,cotta} are applied to this VLM-based segmentation framework, the improvements remain limited and can even become negative (see Fig.~\ref{fig:motivation}(d)).
We attribute this phenomenon to the nature of pretrained VLMs. Unlike VFMs, VLMs already learn an aligned image–text representation space during pretraining. Directly fine-tuning VLMs with noisy image–text pseudo-labels may disrupt this alignment, leading to unstable optimization and limited performance improvements.

\begin{figure*}[t]
    \centering
\includegraphics[width=\linewidth]{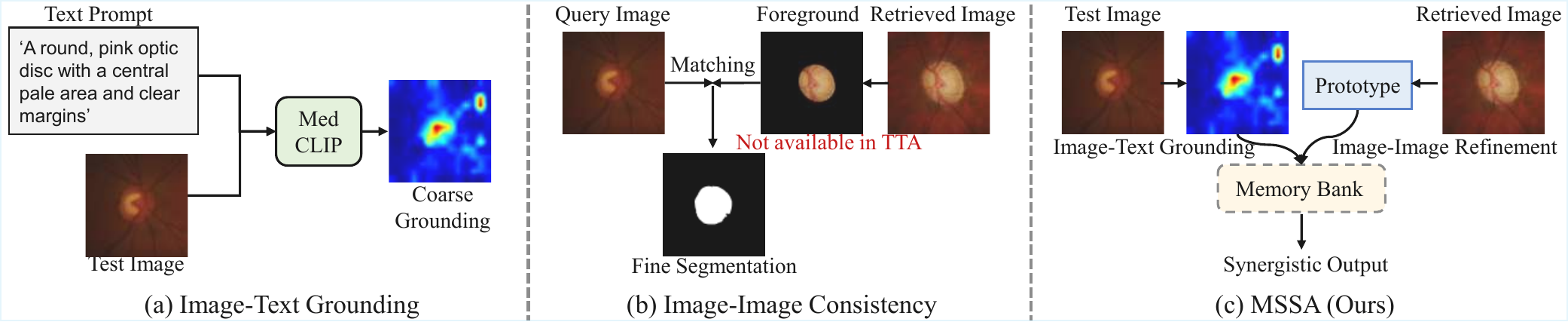}
\captionof{figure}{
Motivation of MSSA. (a) Image--text grounding provides semantic priors for localization but becomes coarse under domain shifts.
(b) Image--image consistency enforces structural alignment, yet lacks semantic anchoring to reliably localize foreground regions in the target domain.
(c) MSSA synergistically combines both, achieving robust and accurate segmentation.}
\label{figit}
\end{figure*}

To overcome these limitations, we propose a novel training-free TTA framework for medical image segmentation, termed Memory-Supported Synergistic Adaptation (MSSA) (see Fig.~\ref{fig:motivation}(c)). 
Instead of updating model parameters with noisy image--text predictions, MSSA transforms them into a support memory that leverages image-to-image consistency to guide the segmentation of all subsequent test images.
Our motivation stems from an intriguing observation illustrated in Fig.~\ref{figit}. 
Image--text grounding in VLMs provides strong semantic priors due to vision--language pretraining, yet it often produces structurally coarse predictions under domain shifts. 
In contrast, image-to-image similarity preserves structural consistency across samples but lacks explicit semantic anchoring to identify target regions. 
Motivated by this complementarity, MSSA integrates semantic grounding with structural consistency, enabling robust adaptation under domain shifts.

Specifically, MSSA maintains an online memory that collects reliable predictions produced by image–text grounding and uses them as structural references for subsequent test images. 
For each incoming query image, MSSA retrieves representative samples from the memory and performs prototype-level image-to-image matching to refine the segmentation prediction.
In this paradigm, we identify two key challenges: how to identify reliable predictions from noisy image–text outputs, and how to effectively utilize the memory to guide adaptation without parameter updates. 
To address the first challenge, we introduce a noise-aware memory construction mechanism that stabilizes cross-modal predictions through multi-augmentation consistency and dual-criterion reliability scoring. 
To address the second challenge, we design a relevance-driven image-to-image adaptation strategy that performs similarity-aware prototype alignment, enabling the memory to provide structured guidance for each query image. 
Together, these designs enable MSSA to convert noisy image–text predictions into structured guidance for training-free TTA, leading to notable improvements over existing TTA methods (e.g., +12.2\% DSC and +11.7\% mIoU in Fig.~\ref{fig:motivation}(d)).

Our contributions are summarized as follows:
\begin{itemize}[label=$\bullet$]
\item We propose the first training-free TTA framework for medical image segmentation, enabling effective adaptation of VLMs at test time while mitigating error accumulation, without introducing sensitive hyperparameter tuning.
\item We introduce a noise-aware memory construction mechanism that identifies reliable image--text predictions via augmentation consistency and reliability scoring, enabling stable memory formation for adaptation.
\item We develop a relevance-driven image-to-image adaptation strategy that retrieves representative samples from memory and performs prototype alignment to guide segmentation without parameter updates.
\item Extensive experiments on multiple medical datasets demonstrate that MSSA achieves state-of-the-art performance and exhibits superior generalization compared to existing TTA methods.
\end{itemize}

\section{Related works}
\label{sec:relatedworks}
\par\noindent\textbf{Foundation Models for Medical Image Segmentation.}
The advent of large-scale VLMs like CLIP~\cite{clip} and segmentation models like SAM~\cite{sam} has spurred their adaptation for medical imaging, aiming for versatile, zero-shot segmentation. Models like BiomedCLIP~\cite{biomedclip} have advanced visual-language alignment in the medical domain, demonstrating strong cross-modal retrieval capabilities. To inject semantic awareness into the promptable but semantic-agnostic SAM, a line of research~\cite{salip,medclipv1,medclipv2}, employs VLMs to generate semantic-aware prompts or attention maps for SAM automatically. However, these methods based on the image-text paradigm often suffer from inconsistent and unstable attention generation. To address the inherent instability, we propose to incorporate the methods based on the image-image paradigm that can provide stability and reliable segmentations into a unified design.

\par\noindent\textbf{Test-Time Adaptation.}
TTA aims to adapt a source-pre-trained model to unlabeled test data in a source-free and online manner~\cite{cotta}. Mainstream TTA methods often rely on self-supervised signals, such as entropy minimization~\cite{tent,eata,li2024robust}) or consistency regularization with pseudo-labels~\cite{cotta,zhao_pami_2025}, and frequently update model parameters, e.g., via Batch Normalization layers~\cite{dua}. While effective for classification, these strategies are notoriously vulnerable in dense prediction tasks like segmentation, where errors in pseudo-labels can accumulate and lead to catastrophic performance collapse~\cite{vptta}. This risk is exacerbated in a fully source-free medical setting, where no ground truth is available. Unlike finetuning-based approaches, which may distort pretrained features, our work explores a training-free adaptation strategy that leverages foundational models directly, circumventing the peril of irreversible error propagation.

\par\noindent\textbf{Cross-Image Consistency.}
The principle of cross-image semantic consistency, where shared categories exhibit invariant feature representations, has been widely exploited to enhance segmentation robustness~\cite{i2i,i2i2,persam}. In supervised and few-shot learning, prototype-matching~\cite{fewshot,protosam,persam,Matcher} leverages labeled support sets to resolve local ambiguities. In semi-supervised segmentation, Querying Labeled for Unlabeled~\cite{i2i2} utilizes this consistency to propagate knowledge from reliable anchors to unlabeled samples. These works collectively demonstrate that inter-image relationships can effectively mitigate individual instance errors.
However, existing consistency-based frameworks fundamentally rely on ground-truth labels or source-domain priors, limiting their utility in TTA. To circumvent the need for manual labels, we construct a high-quality memory bank autonomously curated from the outputs of image-text foundational models.

\begin{figure*}[t]
    \centering
    \includegraphics[width=0.9\linewidth]{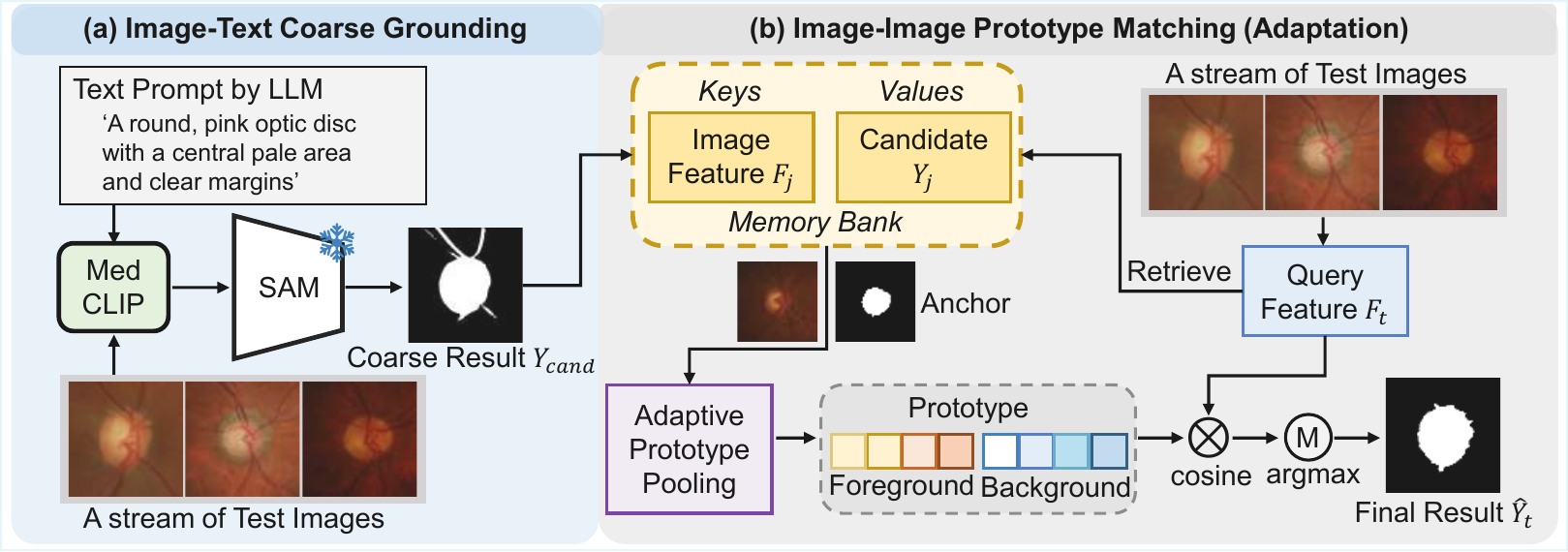}
    \caption{The overall framework of MSSA operating in two stages. \textbf{(a)} Image-Text Coarse Grounding produces initial coarse masks from input images and text prompts, stabilized via our dual majority voting. \textbf{(b)} Noise-Aware Memory Construction filters these candidates by evaluating their semantic alignment and spatial smoothness, populating a dynamic memory bank with only the highest-quality samples. Image-Image Prototype Matching retrieves the most similar anchor example for a query image to compute a prototype and generate the final, accurate segmentation.
}
    \label{fig:framework}
\end{figure*}
\section{Methods}
\noindent\textbf{Task Definition.} We address the problem of TTA for medical image segmentation. Given a stream of unlabeled test images $\{I^i\}_{i=1}^N$ from a target domain, our goal is to predict their corresponding segmentation masks $\{\hat{Y}_t^i\}_{i=1}^N$. The objective is to adapt to the target domain without access to any source data or ground-truth labels, relying solely on pre-trained source or foundation models and continuously arriving target samples.

\noindent\textbf{Method Overview.} As illustrated in Fig.~\ref{fig:framework}, our training-free framework achieves the above objective through a synergistic, two-stage pipeline without any parameter updates or backpropagation: (1) Image-Text Coarse Grounding for producing high-quality initial predictions, (2) Adaptation: Noise-Aware Memory Construction for filtering and retaining reliable samples; and Image-Image Prototype Matching for robust and consistent segmentation. 

\subsection{Image-Text Coarse Grounding}
This stage aims to produce initial, class-aware segmentations by leveraging the zero-shot capability of BiomedCLIP~\cite{biomedclip}. We build upon MedCLIP-SAMv2~\cite{medclipv2} as the base pipeline, and propose several key improvements to further enhance its robustness, as shown in Fig~\ref{fig:framework} (a).

\noindent\textbf{Base Pipeline.} Given a test target image $I$, we utilize a descriptive text prompt $T$ (generated via an LLM GPT-4~\cite{gpt4}) and the BiomedCLIP model to obtain a saliency map $M_s$ through the M2IB module~\cite{M2IB}. The coarse saliency map is then post-processed into a binary mask $M$, which serves to generate the visual prompts (points and bounding boxes) for SAM to obtain a zero-shot segmentation prediction $Y_{cand}$, following~\cite{medclipv2}. Details are shown in the supplementary.

\noindent\textbf{Gaussian Point Selection.} A key limitation of the standard pipeline is the reliance on random point prompts within the coarse mask $M$, which can be sensitive to boundary inaccuracies. Instead, we propose a {Gaussian Point Selection} strategy. We compute the distance transform of the coarse mask and sample points with a probability weighted by a Gaussian kernel centered at the mask centroid $(x_0, y_0)$:
\begin{equation}
GM(x,y) = \exp\left(-\frac{(x-x_0)^2 + (y-y_0)^2}{2\sigma^2}\right),
\end{equation}
where $\sigma$ is the scale parameter controlling the spatial spread of the Gaussian distribution. This prioritizes points near the object center, which are more stable and reliable, leading to more consistent refinements by SAM.

\noindent\textbf{Stabilization via Dual Majority Voting.} To counteract the instability of VLM-generated saliency maps, we employ a dual majority voting scheme.

For Saliency Maps: We apply photometric augmentations (e.g., CLAHE, Gamma Correction) to the input image, generate a saliency map for each variant, and aggregate them via majority voting to produce a consolidated, more stable saliency-based mask $M$.

For SAM Refinement: We apply geometric augmentations (e.g., flips, scales, rotations) to the image-prompt input pair of SAM and aggregate the corresponding SAM outputs via majority voting. This enforces geometric consistency and yields the candidate segmentation $Y_{cand}$.

\subsection{Noise-Aware Memory Construction}
The coarse candidate segmentations $\{Y_{cand}\}$ generated from the previous stage exhibit high potential but are unstable for direct use. 
To exploit their reliability while suppressing noise, we employ a selective scoring procedure to construct a high-quality \textbf{Memory Bank}, denoted as $\mathcal{B} = \{(F_j, Y_j)\}$.

It stores pairs of image features $F_j$ and their corresponding segmented masks $Y_j$. To determine which candidates should be retained, we introduce a dual-faceted scoring mechanism that jointly evaluates two complementary aspects of segmentation quality:
(1) semantic alignment score $\mathcal{Q}_{sem}$ with the target textual concept, and
(2) spatial smoothness score $\mathcal{Q}_{smo}$ that reflects boundary consistency.
The overall score $\mathcal{Q}_{total}$ for each candidate is defined as:
\begin{equation}
\mathcal{Q}_{total} = \mathcal{Q}_{sem} + \mathcal{Q}_{smo},
\end{equation}

\noindent\textbf{Semantic Alignment Score ($\mathcal{Q}{sem}$).} This score quantifies the congruence between the segmented region and the textual concept. Given the candidate mask $Y_{cand}$, we extract the foreground region from the original image $I$ to obtain a masked image $I_{masked}$. We then compute the BiomedCLIP image-text similarity between $I_{masked}$ and the original text prompts $T$:
\begin{equation}
\mathcal{Q}_{sem} = \text{BiomedCLIP}(I_{masked}, T).
\end{equation}
A high score indicates that the masked region is semantically well-aligned with the target class description.

\noindent\textbf{Spatial Smoothness Score ($\mathcal{Q}_{smo}$).} This score evaluates the structural integrity and boundary smoothness. 
The computation process is summarized in Algorithm~\ref{alg:spatial_smoothness}, which integrates global shape regularity and local boundary stability through two key components:
(1) Shape Regularity ($R_s$): This metric, derived from the area $A(C)$ and perimeter $P(C)$, favors shapes that are spatially consolidated. It ensures the mask resembles a well-formed anatomical structure rather than fragmented noise, without over-penalizing natural biological variations.
(2) Boundary Smoothness ($B_s$): Calculated from the vertex density of the approximated polygon $C_{approx}$, this term focuses on local contour stability. Fewer high-curvature vertices imply a smoother contour, effectively reducing artifacts and noise-induced irregularities.

\begin{algorithm}[htbp]
\caption{Spatial Smoothness Score Computation}
\label{alg:spatial_smoothness}
\begin{algorithmic}[1]
\item[] \textbf{Input:} Binary mask $Y_{cand}$
\item[] \textbf{Output:} Spatial smoothness score $\mathcal{Q}_{smo}$
\State \textbf{Step 1: Contour Extraction \& Approximation}
\State Extract the maximum external contour $C$ from $Y_{cand}$
\State Approximate $C$ with a polygon $C_{approx}$
\State \textbf{Step 2: Shape Regularity}
\State Compute area $A(C)$ and perimeter $P(C)$
\State Calculate shape regularity score $R_s$
\State \textbf{Step 3: Boundary Smoothness}
\State Compute vertex number $N_v$ of $C_{approx}$
\State Calculate boundary smoothness score $B_s$
\State \textbf{Step 4: Score Fusion}
\State $\mathcal{Q}_{smo} \gets R_s + B_s$
\State \Return $\mathcal{Q}_{smo}$
\end{algorithmic}
\end{algorithm}


\noindent\textbf{Adaptive Candidate Selection.}
Candidates with a total score $\mathcal{Q}_{total}$ exceeding a dynamic threshold $\tau_t$ are regarded as \textit{promising} and admitted into the memory bank $\mathcal{B}$ under a FIFO (First-In-First-Out) update policy:
\begin{equation}
    \mathcal{Q}_{total}^t \ge \tau_t.
\end{equation}

Instead of relying on a fixed, dataset-specific threshold which may be sub-optimal under varying domain shifts, we introduce an adaptive thresholding mechanism that automatically calibrates the selection criterion according to the evolving score distribution of the target stream. Specifically, we maintain a historical score buffer
\begin{equation}
\mathcal{H}_t = \{ \mathcal{Q}_{total}^i \}_{i=1}^{t},
\end{equation}
which records the total scores of all processed samples up to time $t$. 
After a short warm-up period of $N_{warm}$ samples ($N_{warm}=10$), the threshold is updated:

\begin{equation}
\tau_t =
\begin{cases}
\tau_{\text{init}}, & t < N_{warm}, \\
\max\!\left(\tau_{t-1}, \operatorname{Percentile}(\mathcal{H}_t, P)\right), & t \ge N_{warm},
\end{cases}
\end{equation}
where $P$ denotes a high percentile (i.e., $P=80$).

This percentile-based relative thresholding removes dependence on absolute score magnitudes and enables the model to adaptively align its selection strictness with the intrinsic difficulty of the target domain. 

Importantly, we enforce a \textit{monotonically non-decreasing} update rule to prevent degradation of the memory bank quality over time. Without this constraint, temporary fluctuations in score distribution could lower the threshold and admit noisy candidates, leading to error accumulation in subsequent prototype matching. The monotonic design ensures that once high-quality samples are identified, the selection criterion will not relax, thereby maintaining a progressively refined and high-fidelity memory.
In practice, the adaptive percentile-based update rapidly supersedes the initialization after the warm-up phase, rendering the framework largely insensitive to the specific choice of $\tau_{\text{init}}$. 

Each selected candidate $Y_j$ is stored in $\mathcal{B}$ as the value, where the key is the corresponding DINOv2~\cite{dinov2} image feature $F_j$, as in Fig~\ref{fig:framework} (b).

\noindent\textbf{Remark}: This curation process ensures that $\mathcal{B}$ contains only semantically precise and spatially clean examples, forming a robust foundation for the subsequent prototype matching stage. Noise-Aware Memory Construction (NMC) enables the possibility of harnessing the high ceiling of the image-text modality while mitigating its inconsistency.

\subsection{Robust Segmentation via Image-Image Prototype Matching}
With a curated Memory Bank $\mathcal{B}$, we are able to segment testing samples using a robust, non-parametric, image-image prototype matching approach, as in Fig~\ref{fig:framework} (b). We treat the current test sample $I$ as the query and extract its DINOv2~\cite{dinov2} feature $F_t$. If $\mathcal{B}$ is not empty, select the anchor pair $(F_a, Y_a)$ that has the highest feature similarity to the test feature $F_t$; otherwise, the current initial segmentation $Y_{cand}$ serves as anchor.

\noindent\textbf{Anchor-Query Matching.} For a test query image $I$, we first identify its most relevant anchor example from the bank. We compute the feature similarity between the test feature $F_t$ and all anchor features $F_j$ in $\mathcal{B}$:
\begin{equation}
s_j = \frac{F_t \cdot F_j}{\|F_t\| \|F_j\|}, \quad (F_j, Y_j) \in \mathcal{B}.
\end{equation}
We select the anchor pair $(F_a, Y_a)$ with the highest similarity $s_j$ to guide the segmentation of $I$.

\noindent\textbf{Non-Parametric Prototype Extraction.} We adapt the ALPNet~\cite{alpnet} into our framework. Using the selected anchor mask $Y_a$, we extract features from the corresponding anchor feature map $F_a$ for both foreground ($c=1$) and background ($c=0$) classes. Specifically, we compute a set of local prototypes by applying average pooling with a sliding window over the regions defined by $Y_a$:
\begin{equation}
P_c^{(m,n)} = \frac{1}{| \Omega^{(m,n)} |} \sum_{u,v \in \Omega^{(m,n)}} F_a(u,v) \cdot f_{\theta}[Y_a(u,v) = c].
\end{equation}
where $\Omega^{(m,n)}$ is the local window centered at $(m,n)$, and $f_{\theta}$ is the indicator function. This yields prototype sets $\mathcal{P}_1$ and $\mathcal{P}_0$ for foreground and background, respectively.

\noindent\textbf{Similarity Computation \& Mask Decoding.} The test segmentation is generated by comparing its image feature $F_t$ against the extracted prototypes. For each prototype $P_l^c \in \mathcal{P}_c$, we compute a cosine similarity map:
\begin{equation}
M_l^c(u,v) = \frac{F_t(u,v) \cdot P_l^c}{\|F_t(u,v)\| \|P_l^c\|}.
\end{equation}
These maps are aggregated into a class-wise similarity map by taking the element-wise maximum for each class: $\tilde{M}^c = \max_l M_l^c$. The final segmentation logits $L$ are obtained by normalizing the foreground and background similarity maps and predicted test mask $\hat{Y}_t$ is derived from $L$:
\begin{equation}
L = \text{softmax}([\tilde{M}^1; \tilde{M}^0]), \hat{Y}_t = \arg\max L.
\end{equation}
The detailed algorithm is in the supplementary.

\noindent\textbf{Remark}: Crucially, the entire process is training-free. We do not fine-tune any models or parameters. Instead, we rely on the powerful representations of the foundation model and the similarity matching provided by our NMC. This design enables robust image-to-image matching without any ground-truth supervision.


\section{Experiments}

\subsection{Datasets and Evaluation Metrics}
Following TTCS~\cite{TTCS}, we evaluated our method under TTA setting on two medical segmentation tasks: optic disc (OD) segmentation in fundus images and lung segmentation in chest X-rays. 
{Optic Disc \& Cup:} We collected five public fundus image datasets, treated as distinct domains. The domains include: A (RIM-ONE-r3)~\cite{RIM} with 159 images, B (REFUGE)~\cite{Refuge} with 400 images, C (ORIGA)~\cite{origa} with 650 images, D (REFUGE-Valid)~\cite{Refuge} with 800 images, and E (Drishti-GS)~\cite{Drishti} with 101 images. All images were cropped to a $1024\times1024$ region of interest centered on OD. 
{Chest X-ray Lung Segmentation:} To evaluate adaptation across diverse anatomical and pathological conditions, we utilized two benchmark chest X-ray datasets. First, the MC \& SZ dataset~\cite{MCSZ} served as a cross-site benchmark, comprising 138 images from a US tuberculosis program and 566 from Shenzhen Hospital, China. To assess scalability and robustness, we employed the COVID-QU-Ex database~\cite{lungxray1,lungxray2}, specifically selecting the 957 test images covering normal, lung opacity, viral pneumonia, and COVID-19 cases, as the target domain. This diverse selection tested the framework's stability against both institutional domain shifts and significant pathological variations.
Following previous works~\cite{TTCS,opsam}, we employed Dice Score Coefficient (DSC) and Mean Intersection over Union (mIoU) for evaluation.

\subsection{Implementation Details}
Following the protocols established in~\cite{TTCS, medclipv2}, we conducted our segmentation TTA experiments using a batch size of 1 to ensure evaluation consistency. We employed SAM~\cite{sam}, BioMedCLIP~\cite{medclipv1}, and DINOv2~\cite{dinov2} as our foundational models. All images were resized to $1024 \times 1024$ to meet SAM’s input requirements. For data augmentation, we applied CLAHE and gamma correction~\cite{gamma} to generate attention maps, while random flipping, rotation, and scaling were used to create four augmented variants for SAM. The memory bank is set to store up to 15 samples. For the adaptive thresholding mechanism, we set the parameters to $P=80$ and $N_{warm}=10$, with the hyperparameter $\sigma$ empirically determined as $0.25$.

\subsection{Comparison with State-of-the-art Methods}
We conducted a comprehensive evaluation against existing SOTA methods, which we categorized into two primary groups based on their adaptation strategy: (1) \textbf{Training-based TTA methods}: These approaches adapt model parameters using test-time data. This group included TENT~\cite{tent}, EATA~\cite{eata}, CoTTA~\cite{cotta}, DELTA~\cite{delta}, SVDP~\cite{SVDP}, and TTCS~\cite{TTCS}; and (2) \textbf{Training-free Zero-shot Segmentation methods}: These methods leverage the inherent capabilities of large-scale foundation models without performing any parameter updates or online adaptation. Representative baselines included SaLIP~\cite{salip} and MedCLIPv2~\cite{medclipv2}.


Following the experimental protocol established in TTCS~\cite{TTCS}, we adopted its prompt generation strategy for baseline methods not inherently compatible with SAM to ensure a fair comparison. In this framework, SAM served as the universal segmentation backbone across all evaluated approaches. Initial predictions generated by SAM were utilized as baseline outputs. Subsequently, each method was fine-tuned on these predictions using Low-Rank Adaptation (LoRA)~\cite{lora}, adhering to the specific optimization objectives and adaptation procedures defined in the original implementations.

\begin{table}[h]
\centering
\caption{Quantitative comparison of TTA methods built upon the VLM-based segmentation model MedCLIPv2 on the optic disc dataset. Subscripts denote the relative performance change compared with MedCLIPv2.}
\label{tab:performance-od}
\resizebox{\linewidth}{!}{
\begin{tabular}{c|c|cc|cc|cc|cc|cc|cc}
\hline
\textbf{Technique} & \textbf{Method} & \multicolumn{2}{c|}{\textbf{Domain A}} & \multicolumn{2}{c|}{\textbf{Domain B}} & \multicolumn{2}{c|}{\textbf{Domain C}} & \multicolumn{2}{c|}{\textbf{Domain D}} & \multicolumn{2}{c|}{\textbf{Domain E}} & \multicolumn{2}{c}{\textbf{Average}} \\
\hline
 & & DSC & mIoU & DSC & mIoU & DSC & mIoU & DSC & mIoU & DSC & mIoU & DSC & mIoU \\
\hline
\multirow{5}{*}{VLM}
& SaLIP~\cite{salip} & 42.2 & 35.3 & 30.5 & 21.3 & 27.3 & 19.8 & 14.2 & 5.2 & 35.6 & 26.6 & 30.0 & 21.6 \\
& SAMAug~\cite{samaug} & 52.2 & 49.3 & 73.0 & 66.6 & 53.5 & 37.4 & 69.2 & 63.3 & 61.7 & 54.2 & 61.9 & 54.2 \\
& MedCLIPv1~\cite{medclipv1} & 38.6 & 30.2 & 51.9 & 41.2 & 59.9 & 50.2 & 44.1 & 33.2 & 54.2 & 45.1 & 49.7 & 40.0 \\
& MedCLIPv2~\cite{medclipv2} & 59.8 & 44.5 & 82.2 & 71.0 & 64.9 & 49.8 & 71.1 & 57.2 & 79.0 & 67.2 & 71.4 & 58.0 \\
\hline
\multirow{8}{*}{Training-based TTA}
& TENT~\cite{tent} & 59.8 & 44.6 & 82.1 & 70.8 & 64.9 & 49.8 & 71.4 & 57.5 & 79.6 & 68.2 & 71.6$_{\textcolor{ForestGreen}{(+0.3\%)}}$ & 58.2$_{\textcolor{ForestGreen}{(+0.3\%)}}$ \\
& OCL~\cite{ocl} &59.6&44.3& 82.0&70.8& 64.8&49.7 &70.9&56.9 & 79.6&68.0& 71.4$_{\textcolor{gray}{(+0.0\%)}}$ & 57.9$_{\textcolor{red}{(-0.2\%)}}$ \\
& EATA~\cite{eata} &58.6&43.4& 82.2&71.0& 64.7&49.6 &71.1&57.2& 78.6&66.8&71.0$_{\textcolor{red}{(-0.6\%)}}$ & 57.6$_{\textcolor{red}{(-0.7\%)}}$ \\
& CoTTA~\cite{cotta} &59.3&44.0 &82.0&70.7& 65.0&49.9 & 71.3&57.4  & 77.4&65.7  &71.0$_{\textcolor{red}{(-0.6\%)}}$ & 57.5$_{\textcolor{red}{(-0.9\%)}}$ \\
& SVDP~\cite{SVDP} & 59.4 &44.0 & 82.3 &71.1 & 64.9&49.8& 71.2 &57.2  & 79.0 &67.4  & 71.4$_{\textcolor{gray}{(+0.0\%)}}$ & 57.9$_{\textcolor{red}{(-0.2\%)}}$ \\
& UniVPT~\cite{univpt} & 63.5&48.3 &74.3&60.0& 63.7&48.1 &66.4&50.7& 67.8&52.9 & 67.1$_{\textcolor{red}{(-6.0\%)}}$ & 52.0$_{\textcolor{red}{(-10.3\%)}}$ \\
& DELTA~\cite{delta} & 59.1&43.7& 82.1&70.8& 64.8&49.7& 71.2&57.2& 78.2&66.8 & 71.1$_{\textcolor{red}{(-0.4\%)}}$ &57.6$_{\textcolor{red}{(-0.7\%)}}$ \\
& TTCS~\cite{TTCS} & 65.3 & 52.4 & 86.7 &\textbf{79.1} & 65.2 & 54.1 & 64.7 & 54.0 & 76.4 & 67.4 & 71.7$_{\textcolor{ForestGreen}{(+0.4\%)}}$ & 61.4$_{\textcolor{ForestGreen}{(+5.9\%)}}$ \\
\hline
Training-free TTA& \textbf{MSSA(Ours)} & \textbf{76.6} &\textbf{63.1}& \textbf{87.0}& 77.1& \textbf{84.6}&\textbf{73.8}& \textbf{81.5}&\textbf{69.5}& \textbf{89.9}&\textbf{82.1}& \textbf{83.9}$_{\textcolor{ForestGreen}{\textbf{(+17.5\%)}}}$ &\textbf{73.1}$_{\textcolor{ForestGreen}{\textbf{(+26.0\%)}}}$\\
\hline
\end{tabular}
}
\end{table}

\par\noindent
\textbf{Quantitative Results on Optic Disc/Cup Segmentation.} As shown in Table~\ref{tab:performance-od}, our method consistently outperforms all competitors across all five domains. Notably, we achieve an average DSC of 83.9\% and IoU of 73.1\%, surpassing the SOTA training-based TTA method, TTCS~\cite{TTCS}, by 12.2\% in DSC and the strongest zero-shot method, MedCLIPSAMv2~\cite{medclipv2}, by 12.5\%. This demonstrates that our framework successfully exploits the powerful zero-shot capabilities of large foundation models while effectively leveraging test-time target knowledge through training-free adaptation.

\begin{table*}[htbp]
\centering
\caption{Quantitative comparison of TTA methods built upon the MedCLIPv2 VLM segmentation baseline on different lung datasets. Subscripts denote relative performance changes with respect to MedCLIPv2.}
\resizebox{\linewidth}{!}{
\begin{tabular}{c|c|cc|cc|cc|cc}
\hline
\textbf{Technique} & \textbf{Method} & \multicolumn{2}{c|}{\textbf{COVID}} & \multicolumn{2}{c|}{\textbf{MC}} & \multicolumn{2}{c|}{\textbf{SZ}} & \multicolumn{2}{c}{\textbf{Average}} \\
\hline
 & & DSC & mIoU & DSC & mIoU & DSC & mIoU & DSC & mIoU \\
\hline

\multirow{5}{*}{VLM}
& SaLIP~\cite{salip} & 63.1 & 56.4 & 62.9 & 45.9 & 66.1 & 49.4 & 64.1 & 53.9 \\
& SAMAug~\cite{samaug} & 47.0 & 40.3 & 65.3 & 53.2 & 53.7 & 47.1 & 55.3 & 46.8 \\
& MedCLIPv1~\cite{medclipv1} & 32.6 & 29.0 & 35.5 & 33.6 & 25.2 & 21.9 & 31.1 & 28.2 \\
& MedCLIPv2~\cite{medclipv2} & 54.6 & 40.7 & 62.1 & 54.6 & 60.2 & 50.5 & 59.0 & 48.6 \\
\hline

\multirow{8}{*}{Training-based TTA}
& TENT~\cite{tent} & 56.3 & 40.0 & 63.5 & 54.3 & 62.4 & 46.1 & 60.7$_{\textcolor{ForestGreen}{(+2.9\%)}}$ & 46.8$_{\textcolor{red}{(-3.7\%)}}$ \\
& OCL~\cite{ocl} & 56.4 & 40.1 & 64.1 & 54.8 & 62.5 & 46.1 & 61.0$_{\textcolor{ForestGreen}{(+3.4\%)}}$ & 47.0$_{\textcolor{red}{(-3.3\%)}}$ \\
& EATA~\cite{eata} & 55.8 & 39.5 & 63.1 & 54.0 & 62.1 & 45.7 & 60.3$_{\textcolor{ForestGreen}{(+2.2\%)}}$ & 46.4$_{\textcolor{red}{(-4.5\%)}}$ \\
& CoTTA~\cite{cotta} & 56.0 & 39.8 & 63.9 & 54.9 & 62.4 & 45.9 & 60.8$_{\textcolor{ForestGreen}{(+3.1\%)}}$ & 46.9$_{\textcolor{red}{(-3.5\%)}}$ \\
& SVDP~\cite{SVDP} & 56.3 & 40.1 & 64.0 & 54.9 & 62.4 & 46.0 & 60.9$_{\textcolor{ForestGreen}{(+3.2\%)}}$ & 47.0$_{\textcolor{red}{(-3.3\%)}}$ \\
& UniVPT~\cite{univpt} & 20.4 & 11.9 & 29.1 & 17.7 & 13.3 & 7.5 & 20.9$_{\textcolor{red}{(-64.6\%)}}$ & 12.4$_{\textcolor{red}{(-74.5\%)}}$ \\
& DELTA~\cite{delta} & 56.1 & 39.9 & 63.8 & 54.7 & 62.3 & 45.9 & 60.7$_{\textcolor{ForestGreen}{(+2.9\%)}}$ & 46.8$_{\textcolor{red}{(-3.7\%)}}$ \\
& TTCS~\cite{TTCS} & 57.8& 44.6& 65.2& 43.1 &65.2& 43.1& 62.8$_{\textcolor{ForestGreen}{(+6.4\%)}}$ &43.6$_{\textcolor{red}{(-10.3\%)}}$ \\
\hline

Training-free TTA & \textbf{MSSA(Ours)} &
\textbf{73.2} & \textbf{58.7} &
\textbf{79.2} & \textbf{69.4} &
\textbf{82.9} & \textbf{72.1} &
\textbf{78.4}$_{\textcolor{ForestGreen}{(+32.9\%)}}$ &
\textbf{66.7}$_{\textcolor{ForestGreen}{(+37.2\%)}}$ \\
\hline
\end{tabular}
}
\label{tab:performance-lung}
\end{table*}

\par\noindent
\textbf{Quantitative Results on Lung Segmentation.} The superiority of our approach is further validated on different chest X-ray datasets (Table~\ref{tab:performance-lung}). We observe a significant performance gain, especially on the challenging COVID-QU-Ex dataset, where we improve the DSC by over 15\% to TTCS. This highlights our method's robustness to substantial domain shifts and pathological variations, a scenario where pure TTA methods often fail due to noise propagation and where zero-shot methods lack specificity.

\begin{figure}[htbp]
    \centering
    \includegraphics[width=0.9\linewidth]{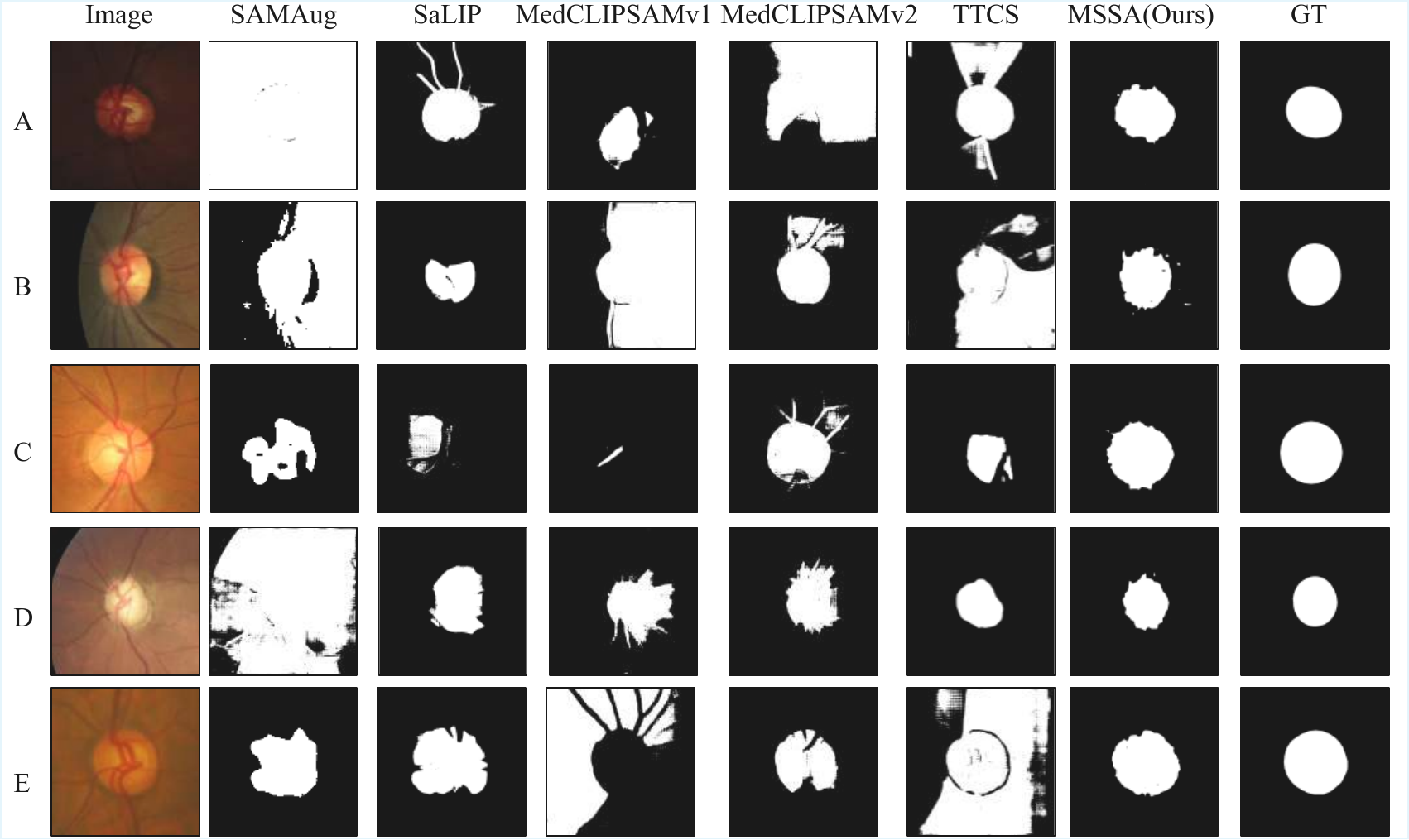}
    \caption{Qualitative comparison of MSSA against state-of-the-art methods on optic disc datasets.}
    \label{fig3}
\end{figure}

\par
\begin{figure*}[htbp]
    \centering
    \includegraphics[width=0.95\linewidth]{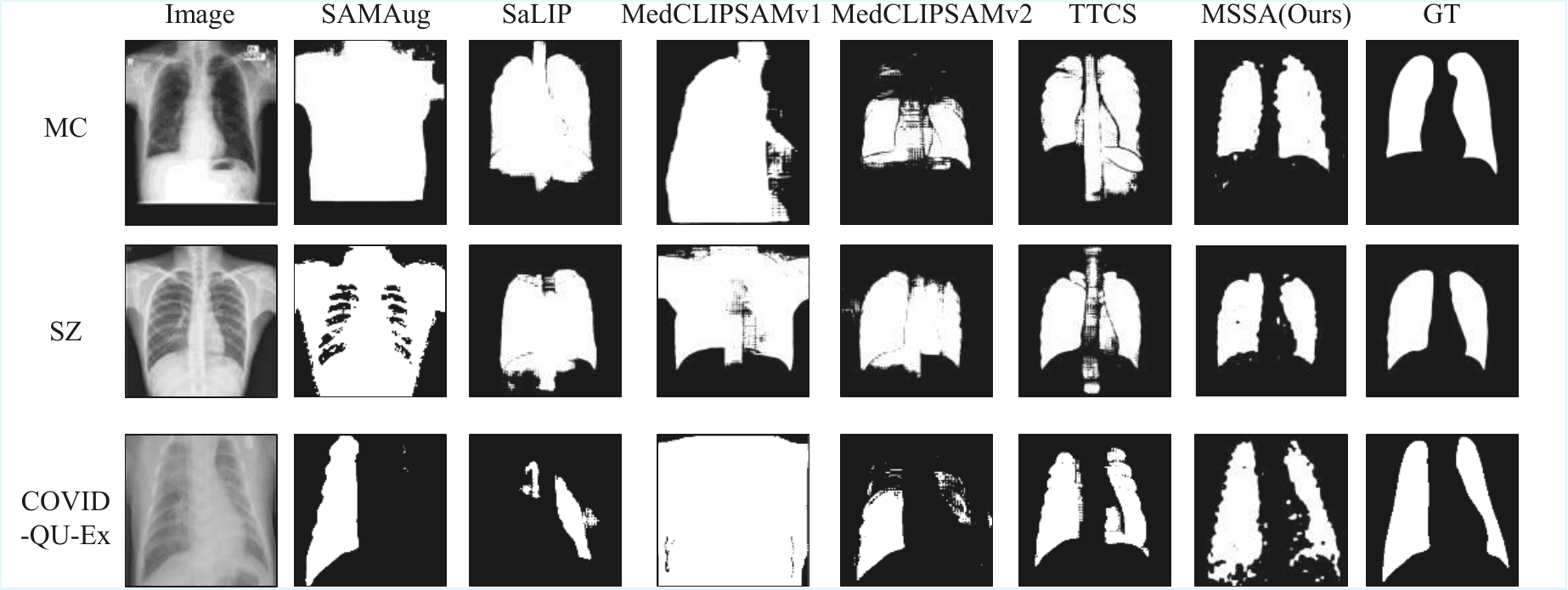}
    \caption{Qualitative comparison of MSSA against other TTA methods on lung datasets.}
    \label{fig4}
\end{figure*}

\noindent
\textbf{Qualitative Results.} Fig.~\ref{fig3} and Fig.~\ref{fig4} provide visual comparisons. The baseline zero-shot methods (e.g., MedCLIPv2) often produce incomplete or semantically inaccurate segmentations. TTA methods (e.g., TTCS), while sometimes more stable, can introduce smoothing artifacts or propagate errors. In contrast, our method generates segmentation masks that are both semantically precise and anatomically coherent, closely aligning with the ground truth. This visually confirms the effectiveness of our selective memory bank in providing stable and accurate guidance.

\subsection{Ablation Study}
We performed extensive ablation studies to verify the contribution of each component in our framework. 


\begin{table*}[htbp]
    \centering
    \caption{Ablation study of key MSSA components on Optic Disc segmentation task. We report mIoU for each target domain and DSC/mIoU averaged over all domains.}
    \setlength{\tabcolsep}{4pt}
    \resizebox{\linewidth}{!}{
    \begin{tabular}{cc|ccc|cc|ccccc|cc}
        \toprule
        \multirow{2}{*}{\textbf{\#}} &
        \multirow{2}{*}{\textbf{Components}} &
        \multicolumn{3}{c|}{\textbf{Image-text}} &
        \multicolumn{2}{c|}{\textbf{Image-Image}} &
        \multicolumn{5}{c|}{\textbf{Domain (mIoU)}} &
        \multicolumn{2}{c}{\textbf{Avg}} \\
        \cmidrule(lr){3-5} \cmidrule(lr){6-7} \cmidrule(lr){8-12} \cmidrule(lr){13-14}
        & & \textbf{SAM-MV} & \textbf{ATN-MV} & \textbf{GP} &
        \textbf{NMC} & \textbf{Similar} &
        \textbf{A} & \textbf{B} & \textbf{C} & \textbf{D} & \textbf{E} &
        \textbf{DSC} & \textbf{mIoU} \\
        \midrule

1 & Baseline & & & & & &
44.5 & 71.0 & 49.8 & 57.2 & 67.2 &
71.4 & 58.0 \\
\midrule[0.4pt]

2 & +SAM-MV & \checkmark & & & & &
47.4 & 76.3 & 53.0 & 56.5 & 75.8 &
72.8 & 61.8 \\
3 & +ATN-MV & \checkmark & \checkmark & & & &
48.8 & 76.3 & 54.9 & 53.3 & 81.0 &
73.9 & 62.9 \\
4 & +GP & \checkmark & \checkmark & \checkmark & & &
51.2 & 76.5 & 57.1 & 57.7 & 81.0 &
75.7 & 64.7 \\
\midrule[0.4pt]

5 & +Bank & & & & \checkmark & &
53.0 & 67.6 & 74.1 & 66.4 & 73.9 &
79.2 & 67.0 \\
6 & +Select & & & & \checkmark & \checkmark &
58.3 & 70.7 & 74.1 & 18.6 & 76.5 &
71.2 & 59.7 \\
\midrule[0.4pt]

7 & All w/o Select & \checkmark & \checkmark & \checkmark & \checkmark & &
63.1 & 76.2 & 73.8 & 67.5 & 73.7 &
83.2 & 70.8 \\
\rowcolor[rgb]{0.8, 0.9, 0.8} 8 & \textbf{Full Model} & \checkmark & \checkmark & \checkmark & \checkmark & \checkmark &
\textbf{63.1} & \textbf{77.1} & \textbf{73.8} & \textbf{69.5} & \textbf{82.1} &
\textbf{83.9} & \textbf{73.1} \\
        \bottomrule
    \end{tabular}}
    \label{tab:image-text-ablation-study}
\end{table*}

\par
\begin{figure}[!t]
    \centering
    \includegraphics[width=0.78\linewidth]{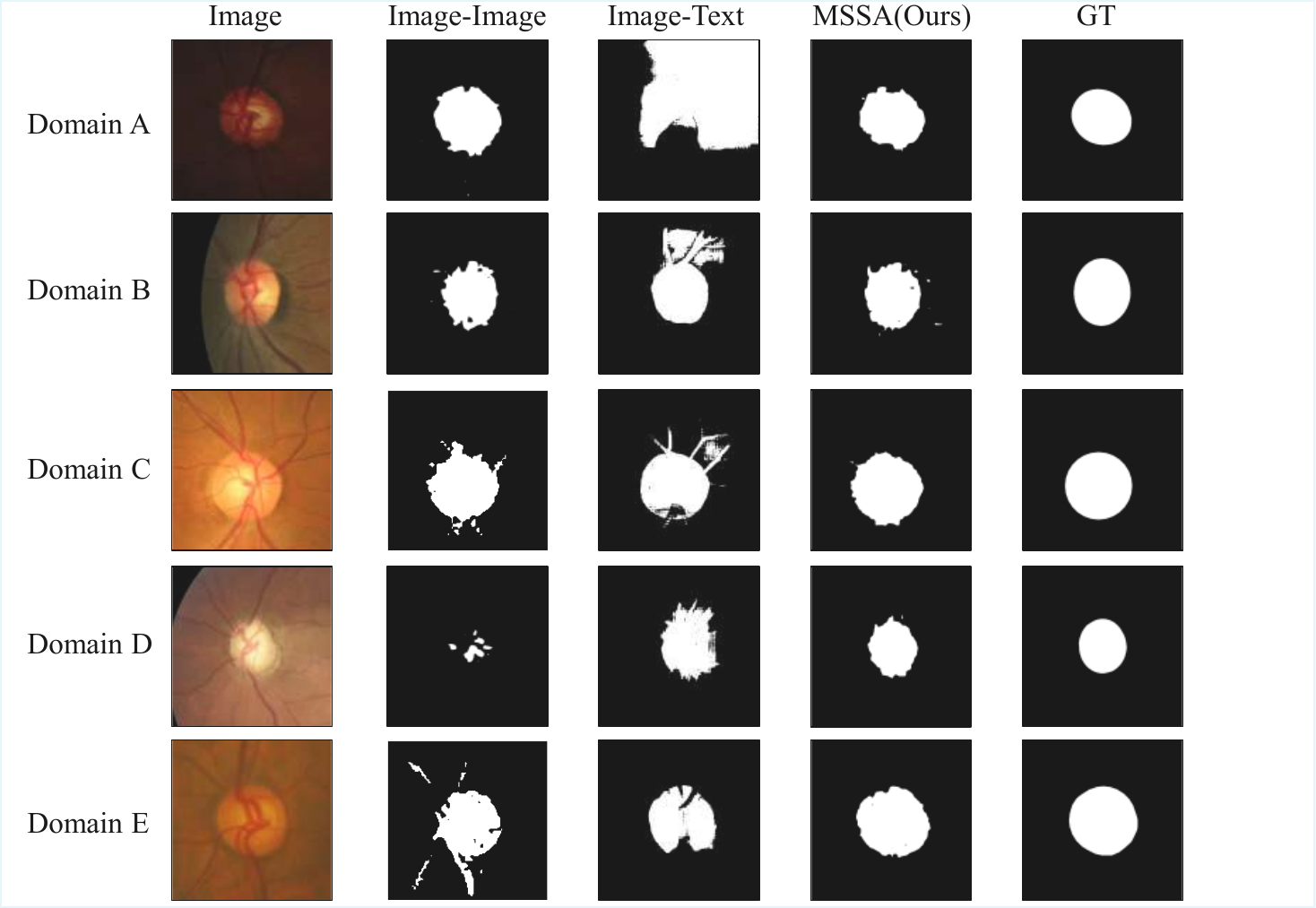}
    \caption{Qualitative comparison of different paradigms on Optic Disc.}
    \label{fig-imgonly1}
\end{figure}

\noindent
\textbf{Component-wise Analysis.} The progression of results in Table~\ref{tab:image-text-ablation-study} (rows 1 to 8) systematically validates our design choices. The experimental conclusions are:
\begin{itemize}[label=$\bullet$]
    \item {Image-Text Branch Stabilization (Rows 2-4):} The introduction of SAM Majority Voting (SAM-MV) and Attention Map Majority Voting (ATN-MV) progressively improves performance, confirming their necessity for stabilizing the initial VLM outputs. Gaussian Point (GP) Selection further provides a noticeable boost, underscoring the importance of robust prompt engineering for SAM.
    \item {Image-Image Branch (Rows 5 \& 6):}  To enable image matching in a zero-shot setting, this image-image variant is initialized with a single image-text sample as a warm-up support, and subsequently relies on its own predictions as the anchor. The results reveal a clear instability in this self-referential strategy, with performance degrading notably over adaptation, particularly causing a catastrophic failure on Domain D, even with NMC. This starkly illustrates the {risk of error accumulation} in a naive self-iterative paradigm.
    \item {The Power of Selective Construction (Rows 4, 7 \& 8):} The critical step, our proposed Noise-Aware Memory Construction, leads to better performance across all domains. This proves that {selecting high-quality candidates is paramount} to enabling robust image-image prototype matching. Row 8 validated further improvement of selecting the most similar one as a anchor sample.
\end{itemize}
Fig.~\ref{fig-imgonly1} also demonstrates the qualitative comparison of different paradigms on OD datasets, where relying solely on one paradigm leads to suboptimal results.

\par\noindent
\textbf{Relevant Anchor Sample Strategy.} We also ablated the criteria for selecting the anchor sample for a given query in Table~\ref{tab:odrandomorsimilar}. Simply choosing a Random sample from the bank or selecting based solely on Semantic Alignment score like $Q_{sem}$ improves but is suboptimal. Our final strategy, which selects the most \textit{Similar} sample based on DINOv2~\cite{dinov2} feature similarity, achieves the best results. This indicates that feature-level similarity is a more reliable indicator for successful prototype transfer than the semantic score alone, as it captures visual and structural affinity.

\begin{table*}[!htbp]
\centering
\caption{Ablation studies of different anchor sample scoring mechanisms for optic disc segmentation.}
\scalebox{0.82}{
\begin{tabular}{c|cc|cc|cc|cc|cc|cc}
\hline
\textbf{Selection} & \multicolumn{2}{c|}{\textbf{A}} & \multicolumn{2}{c|}{\textbf{B}} & \multicolumn{2}{c}{\textbf{C}} & \multicolumn{2}{c}{\textbf{D}}  & \multicolumn{2}{c}{\textbf{E}}   & \multicolumn{2}{c}{\textbf{Avg}}    \\
\hline
 & DSC & mIoU & DSC & mIoU & DSC & mIoU & DSC & mIoU& DSC & mIoU  & DSC & mIoU \\
\hline
Random    & 76.6 & 63.1 & 86.3 & 76.2 & 83.9 & 73.8 & 79.4 & 67.5 & 89.9 & 73.7 & 83.2 & 70.8 \\
Alignment & 76.6 & 63.1 & 85.3 & 74.7 & 82.0 & 70.9 & 80.2 & 68.5 & 89.9 & 73.7 & 82.8 & 70.2 \\
Similar   & \textbf{76.6} & \textbf{63.1} & \textbf{87.0} & \textbf{76.7} & \textbf{84.6}&\textbf{73.8}& \textbf{81.5}&\textbf{69.5}& \textbf{89.9}&\textbf{82.1}& \textbf{83.9} &\textbf{73.1} \\
\hline
\end{tabular}
}
\label{tab:odrandomorsimilar}
\end{table*}

\par\noindent
\textbf{Memory Bank Scoring Mechanism.} We ablated the components of our selective scoring mechanism for memory bank Construction. Table~\ref{tab:ablation-bankselectscore} compares using only the semantic alignment score ($Q_{sem}$), only the spatial smoothness score ($Q_{smo}$), and our combined approach ($Q_{total}$). 
The results demonstrate that neither score alone is sufficient. Relying solely on \textit{Semantic Alignment} can select masks that are semantically correct but spatially fragmented, limiting their effectiveness as anchor examples. Conversely, using only \textit{Spatial Smoothness} may select well-shaped masks that are semantically incorrect, providing misleading guidance. 
Our dual-criteria strategy, requiring excellence in both semantic and spatial quality, achieved superior performance across all datasets. This confirms that these criteria are complementary: the semantic score verifies the accuracy of the segmented class, while the smoothness score ensures the structural integrity and reliability of the boundaries. Together, they provide a necessary filter for curating a high-quality, stable memory bank.
\begin{table}[htbp]
\centering
\begin{minipage}{0.48\textwidth}
\centering
\caption{Ablation studies of selection score strategy on lung datasets.}
\scalebox{0.8}{
\begin{tabular}{c|cc|cc|cc|cc}
\hline
\textbf{Score} & \multicolumn{2}{c|}{\textbf{LungXray}} & \multicolumn{2}{c|}{\textbf{MC}} & \multicolumn{2}{c}{\textbf{SZ}}  & \multicolumn{2}{c}{\textbf{Avg}}  \\
\hline
 & DSC & mIoU & DSC & mIoU & DSC & mIoU & DSC & mIoU \\
\hline
$Q_{sem}$   & 69.9 & 55.5 & 70.4 & 55.6 & 68.9 & 53.6 & 69.7 & 54.9 \\
$Q_{smo}$   & 70.2 & 55.5 & 70.4 & 55.6 & 69.0 & 53.6 & 69.9 & 54.9 \\
$Q_{total}$ & \textbf{73.2} & \textbf{58.7} & \textbf{79.2} & \textbf{69.4} & \textbf{82.9} & \textbf{72.1} & \textbf{78.4} & \textbf{66.7} \\
\hline
\end{tabular}
}
\label{tab:ablation-bankselectscore}
\end{minipage}
\hfill
\begin{minipage}{0.48\textwidth}
\centering
\caption{Comparison of inference time per image (seconds) on Drishti-GS.}
\scalebox{0.7}{
\begin{tabular}{l|c|c}
\hline
\textbf{Method} & \textbf{Performance(\textuparrow)}& \textbf{Time(s)(\textdownarrow)}\\
\hline
OCL~\cite{ocl} & 61.0\% &1.7\\
DELTA~\cite{delta} &60.7\%  &1.7 \\
SVDP~\cite{SVDP} & 60.9\% &2.3 \\
TTCS~\cite{TTCS} & 62.8\% &5.9 \\
MSSA(Ours) &78.4\% & 12.1\\
\hline
\end{tabular}}
\label{tab:tta_time_comparison}
\end{minipage}
\end{table}

\subsection{Limitation} 
Table~\ref{tab:tta_time_comparison} compares average per-image inference latency across different TTA methods on a single NVIDIA RTX A6000 GPU. While MSSA exhibits higher latency due to its multi-augmentation voting and memory retrieval, this is a deliberate trade-off to achieve notable gains in segmentation accuracy and stability.
This trade-off is well-justified for non-real-time medical tasks (e.g., diagnostic planning) where accuracy is critical to avoid misdiagnosis. Currently, MSSA is most effective for organs with regular geometries (e.g., lung, optic disc); its performance on highly intricate structures, like micro-vascular networks, remains a subject for future optimization.

\section{Conclusion}
In this paper, we propose MSSA, a novel training-free framework for test-time adaptation in medical image segmentation.
By integrating three cohesive modules, stabilized candidate generation, dynamic noise-aware memory construction, and proto-type-based refinement, MSSA enables consistent and accurate adaptation without any source data or ground-truth labels.
Conceptually simple and effective, MSSA is readily applicable to real-world medical scenarios without labeled data and with prevalent domain shifts, offering a new perspective for test-time adaptation. 
\section*{Acknowledgments}
This work was funded by the National Natural Science Foundation of China ( No. 62572166 \& No. 62402157) and the Fundamental Research Funds for the Central Universities (No. JZ2025HGTB0219).


%
%
\bibliographystyle{splncs04}
\bibliography{main}
\end{document}